# UmUTracker: A versatile MATLAB program for automated particle tracking of 2D light microscopy or 3D digital holography data

Hanqing Zhang[†], Tim Stangner[†], Krister Wiklund[†], Alvaro Rodriguez[†], Magnus Andersson[†,*]

[†]Department of Physics, Umeå University, 901 87 Umeå, Sweden

[*]**Corresponding author:** Magnus Andersson, magnus.andersson@umu.se

## ABSTRACT

We present a versatile and fast MATLAB program (UmUTracker) that automatically detects and tracks particles by analyzing video sequences acquired by either light microscopy or digital in-line holographic microscopy. Our program detects the 2D lateral positions of particles with an algorithm based on the isosceles triangle transform, and reconstructs their 3D axial positions by a fast implementation of the Rayleigh-Sommerfeld model using a radial intensity profile. To validate the accuracy and performance of our program, we first track the 2D position of polystyrene particles using bright field and digital holographic microscopy. Second, we determine the 3D particle position by analyzing synthetic and experimentally acquired holograms. Finally, to highlight the full program features, we profile the microfluidic flow in a 100 μm high flow chamber. This result agrees with computational fluid dynamic simulations. On a regular desktop computer UmUTracker can detect, analyze, and track multiple particles at 5 frames per second for a template size of 201 × 201 in a 1024 × 1024 image. To enhance usability and to make it easy to implement new functions we used object-oriented programming. UmUTracker is suitable for studies related to: particle dynamics, cell localization, colloids and microfluidic flow measurement.

## PROGRAM SUMMARY

Program title: UmUTracker
Catalogue identifier: 4.6, 4.12, 14, 16.4, 16.10,
Program obtainable from: https://sourceforge.net/projects/umutracker/
Licensing provisions: Creative Commons by 4.0 (CC by 4.0)
Distribution format: zip
Programming language: MATLAB
Computer: All computers running the (MathWorks Inc.) version 8.4 (2014b) or higher, with computer vision toolbox should work.
Operating system: Tested on Windows 7 and 10. All operating systems running MATLAB with computer vision toolbox should fulfill necessary requirements.
Classification: 06.30.Gv, 47.80.Jk, 42.30.Wb, 87.64.-t, 07.05.Pj

*Nature of problem:* 3D multi-particle tracking is a common technique in physics, chemistry and biology. However, in terms of accuracy, reliable particle tracking is a challenging task since results depend on sample illumination, particle overlap, motion blur and noise from recording sensors. Additionally, the computational performance is also an issue if, for example, a computationally expensive process is executed, such as axial particle position reconstruction from digital holographic microscopy data. Versatile robust tracking programs handling these concerns and providing a powerful post-processing option are significantly limited.

*Solution method:* UmUTracker is a multi-functional tool to extract particle positions from long video sequences acquired with either light microscopy or digital holographic microscopy. The program provides an easy-to-use graphical user interface (GUI) for both tracking and post-processing that does not require any programming skills to analyze data from particle tracking experiments. UmUTracker first conduct automatic 2D particle detection even under noisy conditions using a novel circle detector based on the isosceles triangle sampling technique with a multi-scale strategy. To reduce the computational load for 3D tracking, it uses an efficient implementation of the Rayleigh-Sommerfeld light propagation model. To analyze and visualize the data, an efficient data analysis step, which can for example show 4D flow visualization using 3D trajectories, is included. Additionally, UmUTracker is easy to modify with user-customized modules due to the object-oriented programing style.

# 1. INTRODUCTION

Reliable automatic particle tracking is a crucial step in the quantitative analysis of microscopic images in several research fields. However, the accuracy of the analysis result depends on two things: the quality of the acquired images, and thus on the imaging system properties; and the performance of the tracking algorithm, which is challenged significantly if the data is noisy. To handle these concerns improvements of both experimental approaches and development of optimized algorithms have been carried out during recent years [1]. For example, by using digital holographic microscopy (DHM) detailed particle positions can be acquired. DHM is an imaging technique that allows three-dimensional (3D) tracking of semi-transparent objects such as cells, bacteria and particles using amplitude and phase information recorded in holograms [2–4]. Since a hologram contains 3D information in a two-dimensional (2D) image, the objects under study do not have to be in focus in contrast to regular microscopy imaging. DHM is therefore a powerful technique for biophysical and colloidal research studies since multiple objects at different height can be investigated simultaneously. With these advantages, particle image velocimetry for flow profiling of microfluidics can be applied using the DHM technique [5,6]. DHM is also used for measuring the shape and trajectories of *E. coli* [7], red blood cells [8], and position tracking of many different cells types [9–11].

Computational efficiency and reliable tracking are the two major challenges for DHM based tracking using long video sequences. To track particle positions with high accuracy, DHM reconstructs the light intensity from a hologram, which is a computationally expensive process. Several studies have improved the reconstruction efficiency by applying hardware acceleration using graphical processing units (GPU) [5,12] or field programmable gate arrays (FPGA) [13]. However, these approaches lack reliable tracking algorithms. Most tracking methods using GPU or FPGA are based on correlation or convolution in a pre-defined region-of-interest (ROI). Within this ROI, particles need to be identified manually, which can be time-consuming if the particle density is high. Additionally, if new objects enter the ROI, they need to be manually identified, which is problematic especially for the analysis of long video sequences. Therefore, an efficient reconstruction algorithm in the axial direction ($z$) with automatic multi-particle identification in the lateral plane ($xy$) is required. Furthermore, tracking in the presence of particle overlap, motion blur, and noise must be addressed for better robustness [14].

We present an automatic program (UmUTracker) implemented in MATLAB for particle detection and tracking. We provide a robust particle detection algorithm that is based on an isosceles triangle transform and a fast implementation of the Rayleigh-Sommerfeld model. The algorithm is optimized to analyze circular diffraction patterns acquired using in-line DHM but can also handle light microscopy data. The program provides 3D trajectories of multiple particles by first locating particle positions in 2D, and thereafter extracting their axial position using the Rayleigh-Sommerfeld model. Moreover, we provide a post-processing interface where trajectories as well as particle velocity profiles can be visualized. We verify the accuracy of our algorithm by determining the spatial coordinates of micro-particles using synthetic images from ray-tracing simulations as well as from experimental data. The results show that our algorithm can handle background noise and detect partially occluded particles without losing precision in tracking. This allows for a more accurate estimation of the local flow profile since it is possible to use a higher particle density in the field of view. As a proof of concept, we characterize the particle flow in a microfluidic chamber for 1 µm polystyrene particles and compare our results to computational fluid dynamic (CFD) simulations. The program is implemented efficiently to handle 201×201 pixel images of particles at average speed of 5 fps with MATLAB on a regular desktop computer.

## 2. EXPERIMENTAL PROCEDURE

### 2.1. Sample Preparation

To carry out DHM experiments, we use polystyrene (PS) mono-size particles (nominal diameter ± standard deviation (SD): 1.040 ± 0.022 µm, Lot No. 15879, Duke Scientific Corp., 4% w/v) dissolved in phosphate buffered saline (PBS, pH 7.4). We set the final bead concentration by diluting the bead stock solution by 1:400.

For experiments without flow, these polystyrene particles are injected into a sandwich shaped chamber, consisting of two cover slips (lower cover slip: no. 1, Knittel Glass, 60x24 mm; upper cover slip: no. 1, Knittel Glass, 20x20 mm) separated by double sticky tape (Scotch, product no. 34-8509-3289-7, 3M). The cell volume is ~ 10 µL. We inject the diluted bead solution with a micro-pipette and seal the chamber by adding vacuum grease (Dow Corning) on both openings to prevent evaporation. Particles will eventually settle down and immobilize to the bottom coverslip. For experiments with flow and a preset shear stress, we use a commercial microfluidic chamber bio-chip (Vena8, Celix) with the same bead solution as presented above. The Vena8 bio-chip has a rectangular flow channel with dimensions of 28 mm x 400 µm x 100 µm (length, width, height). To produce constant and reproducible flow rates, we use a microfluidic syringe pump (Lambda Vit-Fit, LAMBDA Laboratory Instruments, Switzerland), operating in pulling mode.

### 2.2. Digital In-Line Holography Experiments

The in-line version of our DHM setup is built around an Olympus IX70 inverted microscope, normally used for optical tweezers experiments, equipped with an oil-immersion objective (Olympus PlanApo 60x/1.40 Oil, ∞/0.17, Figure 1) [15]. We illuminate the sample using a quasi-monochromatic light emitting diode (LED, M470F1, Thorlabs) operating at 470 nm. The light source is connected via an optical fiber (M28L02, Thorlabs) to an output collimator (PAF-SMA-7-A, Thorlabs) enabling us to illuminate the sample from above with parallel light and plane wave fronts. The prepared sample cell is mounted onto a piezo stage, which can be positioned in three dimensions (*xyz* coordinates in Figure 1) over a range of 100 µm with nanometer accuracy using piezo actuators (P-561.3CD, Physik Instrumente). The image of particles is magnified through an oil-immersion objective (Olympus PlanApo 60x/1.40 Oil, ∞/0.17), and recorded using a high-speed camera (MotionBLITZ EoSens Cube 7, Mikrotron, pixel size of 8x8 µm) operating at a shutter time of 5 ms. Video sequences are acquired by MotionBLITZDirector2 software with an image size of 1696 x 1710 pixels and frame rate of 200 fps. The lateral (*xy*-plane in Figure 1) conversion factor of the microscopy system is 132 ± 2 nm/pixel (mean ± SD). The whole setup is built in a temperature controlled room at 296 ± 1 K to ensure long-term stability and to reduce thermal drift effects.

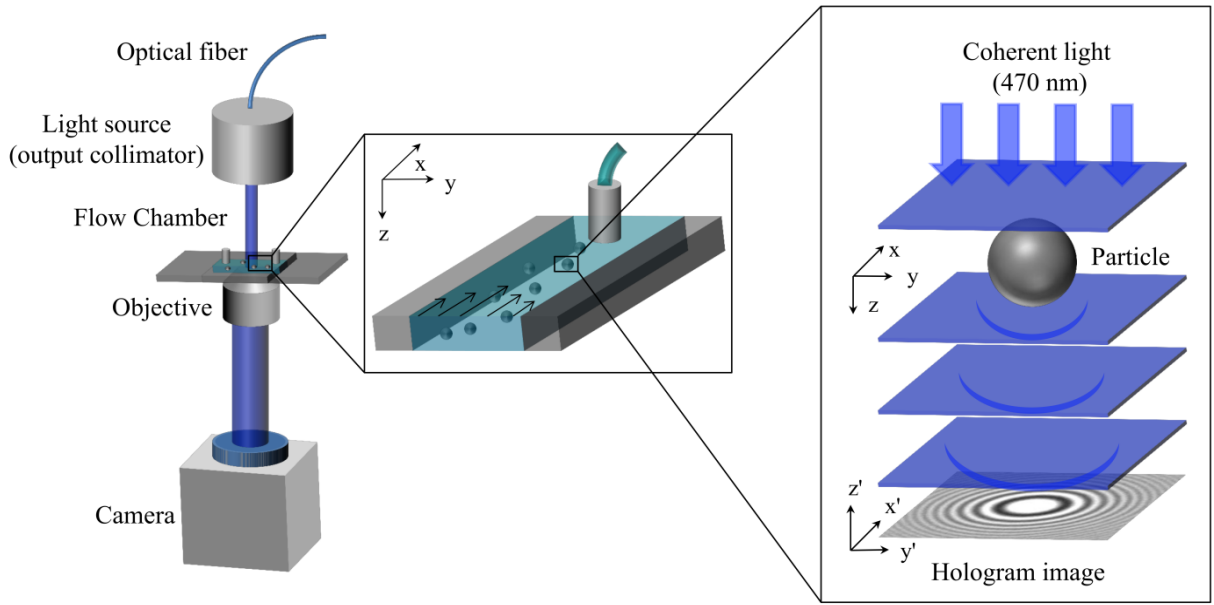

Figure 1. (left) Illustration of the in-line DHM setup. A quasi-monochromatic fiber-coupled LED light source illuminates the sample with plane wave fronts via an output collimator. A flow chamber is mounted on a piezo stage that moves in three dimensions ($xyz$). A hologram is collected by the objective and aquired by a high speed camera ($x'y'z'$). (middle) Schematic representation of particle flow in the flow chamber with polystyrene particles at different heights under constant shear stress. (right) By applying our digital holographic tracking algorithm using holograms, we can recalculate the spatial coordinates of particles with high accuracy.

## 2.3. Bright Field Microscopy Experiments

In addition to analyzing DHM holograms, we also analyzed images of a supercoiled DNA tethered particle acquired in bright field microscopy. The raw data are taken from [16,17]. In the experiment, the supercoiled DNA (pSB4312) was tethered at one end to a coverslip using PNAs while the other end was attached to a 0.5 µm Streptavidin-coated particle (cat. no. CP01N, Bangs Laboratory). The motion of the particle was recorded at 225 Hz in an inverted microscope (Model No. DM IRB, Leica) with a high-speed camera (Pike F100B, Allied Vision Technology, conversion factor of 45.97 nm/pixel), and the length of the time series was optimized to 30 seconds according to an Allan variance analysis of the setup [18].

## 2.4. Computational Fluid Dynamic Simulation

To verify the experimentally recorded flow profiles, we compare our experimental data with computational fluid dynamic (CFD) simulations using the commercial software Comsol Multiphysics 5.2. In these simulations, we simulate a 3D flow in a channel with rectangular geometry of 400 µm x 100 µm, width x height, similar to the Vena8 bio-chip used in the experiments. The channel length is chosen in a way that the fully developed velocity profile is not affected by inlet and outlet conditions and the no-slip velocity condition is used on all walls parallel to the flow.

## 3. ALGORITHM AND PROGRAM DESCRIPTION

### 3.1. Particle Detection using an Isosceles Triangles Transform

Particle tracking using bright field microscopy or DHM requires a pattern recognition algorithm to find the target object. Therefore, finding the center position of multiple particles requires a robust algorithm that can handle occlusions, noise and blurriness in images due to particle motion and lens aberrations. To tackle this issue, we propose a variation of the ITCiD circle detection algorithm where the geometrical constraints of isosceles triangles (ITs) are used to find the circle center [19]. We denoted this as the ITs transform. Compared to the conventional Hough transform for circle detection, the ITs transform is more efficient in suppressing false-positives from noise and finding object centers accurately.

We modify the ITs transform by introducing a coarse-to-fine solution that uses a Gaussian pyramid in the scale-space, which improves both performance and edge detection accuracy of blurry textures in the image [20]. For implementation efficiency, we apply down-sampling to the original image to create multiple images (Figure 2A). At each level of scale, we determine the particle positions from their diffraction pattern in the following way. First, the gradient of the image is calculated and used to find edges of objects. The positions of edges are converted to a binary matrix with the same size as the original image. We then group connected edge points in the binary matrix as line segments and for each segment we apply the ITs transform. The ITs transform produces circle center positions with weight values in the binary matrix. The accumulated weight values at each position in the matrix reveal the probability of potential circle centers. By fusing all the matrixes from all the scales, the algorithm generates the final probability distribution (Figure 2B). This distribution is used to find the pixel position with maximum probability value, corresponding to the $xy$-center coordinates of the diffraction pattern (Figure 2C). Detailed information of ITs implementation can be found in section 3.3.

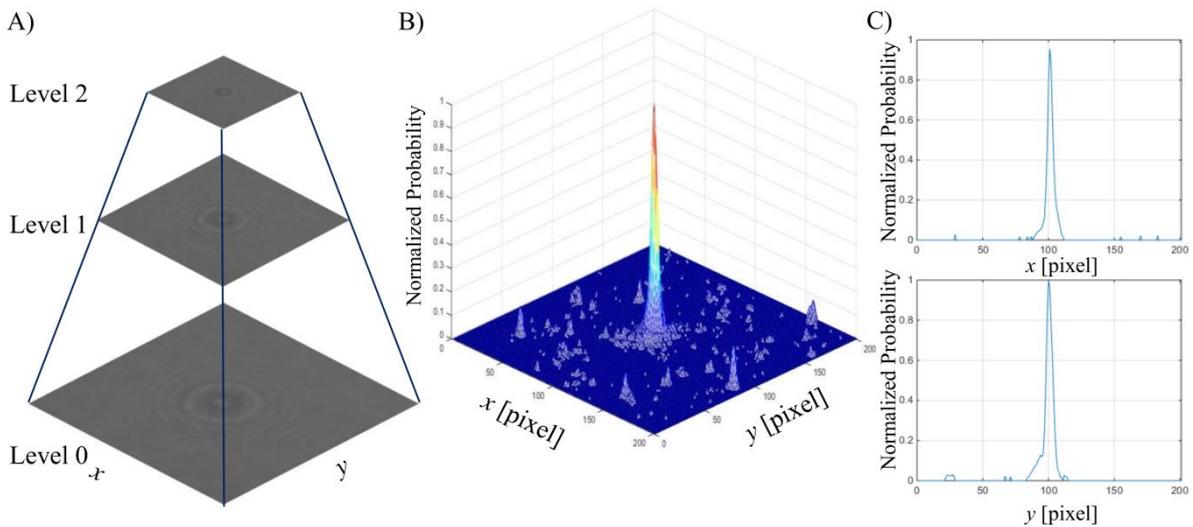

Figure 2. Demonstration of diffraction center detection using the ITs transform. A) Pyramid representation of images containing the original image at level "0" and down sampled images at level "1" and "2". The results of the ITs transform using these images are saved in a probability matrix. B) Fused probability matrix of potential circle centers from each scale level. The matrix is normalized by the maximum probability value. C) 1D slice in $x$ and $y$ from the probability matrix. The center of the diffraction pattern can be found at the point with maximum probability value.

## 3.2. Numerical Reconstruction using Rayleigh-Sommerfeld Back Propagation

When a spherical object is illuminated by monochromatic light with plane wave fronts (Figure 1, right, *z* axis direction in space (*x*,*y*,*z*)), a small proportion of the incident beam is scattered. This scattered light will in turn interfere with non-scattered light and produce an interference pattern. This pattern can be predicted using the Rayleigh-Somerfield diffraction formula for light propagation [21]. The reconstruction of the diffracted field of the particle can be obtained by multiplying a transfer function of free space propagation to the diffraction pattern in the image plane in the frequency domain. The reconstruction can be implemented using the angular spectrum method by [22],

$$U_R(\vec{r}) = FT^{-1}\left\{FT(U_O(\vec{r_0})) \cdot \exp\left[\frac{-2\pi jz}{\lambda}\sqrt{1-\left(\lambda\frac{P}{S}\right)^2-\left(\lambda\frac{Q}{S}\right)^2}\right]\right\}, \qquad (1)$$

where $U_O(\vec{r_0})$ represents the complex optical field scattered in the image plane (*x′*,*y′*,0) (Figure 1, right) and $U_R(\vec{r})$ represents the wave front of the scattered light at any point in space (*x′*,*y′*,*z′*), FT and FT$^{-1}$ are the Fourier transform and the inverse of Fourier transform, *j* is the imaginary unit, $\lambda$ indicates the wavelength of the light used for illumination, S × S is the real size of the hologram, and *P* and *Q* are pixel indexes in *x′* and *y′* dimension respectively ranging from $-N/2$ to $N/2$, where N is the number of pixels.

Since the 3D reconstruction using Equation (1) is time consuming, we propose a fast and robust implementation based on the symmetrical nature of spherical particles. Reconstruction of particle intensity at any height *z* along the center of the particle can be efficiently calculated using a 1D intensity radial profile. To extract the radial profile of particles recorded in the image plane, pre-knowledge of the particle center (as shown in section 3.1) is needed. We resample the intensity values in the image using polar coordinates with their origin at the particle center. In the resampling process, sampling points are spaced by $\Delta$r and $\Delta$q, representing the minimum steps in radial and angular dimensions. At each sampling point, the intensity value is calculated using linear interpolation of two values at neighboring grid points in dimension *x* and *y*, respectively. After resampling, intensities over all angles at each radial step are averaged and the intensity profile $I_O(r', 0)$ of a particle from its center to a certain radius is created. In this process, noise and interference from nearby particles can also be reduced by means of averaging of diffraction pattern in the radial dimension. Finally, a symmetrical radial intensity profile is created by connecting the intensity profile with its mirror.

Subsequently, by using these 1D radial intensity profiles, we numerically reconstruct the axial intensity. Here, we assign axial direction or intensity to the *z* coordinate of the particle center, coinciding with the propagation direction of sample illumination light. Based on Equation (1), the formula for the optical field of the scattered light at any point in 2D space (*r*, *z*), $I_R(r, z)$ is derived as,

$$I_R(r,z) = FT^{-1}\left\{FT(I_O(r',0)) \cdot \exp\left[\frac{-2\pi jz}{\lambda}\sqrt{1-\left(\lambda\frac{R}{S}\right)^2}\right]\right\}, \qquad (2)$$

where *R* is a pixel index ranging from $-M/2$ to $M/2$ and M is the total number of sampled steps. By applying the dimension reduction above, the computational cost for intensity reconstruction of spherical particles can be reduced significantly.

## 3.3. Design of the Framework

The workflow of UmUTracker is presented in Figure 3. All algorithms are implemented in MATLAB using an object-oriented framework. We categorize the abstracted modules as "Process" (blue) and "Mathematics & Method" (grey). Solid lines with arrows indicate the direction of workflow in the process. The dashed lines with arrows show its realization, meaning that the origin of the dependency can be extended or replaced to allow more functionality in the program.

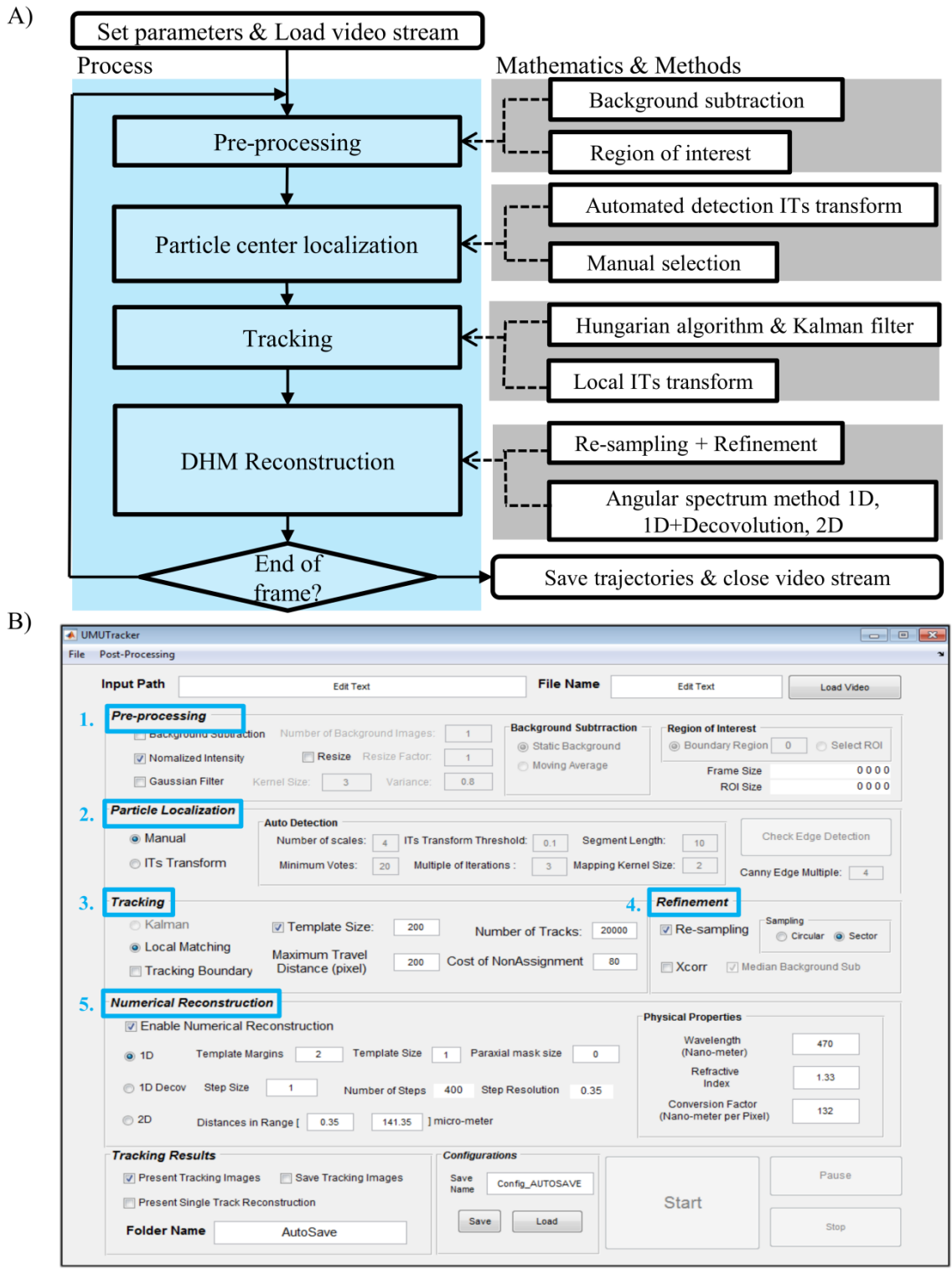

Figure 3. A) Block diagram of the workflow (blue column) and the class modules (grey column) for UmUTracker. B) User interface of UmUTracker containing 5 major functionalities: 1. Pre-processing, 2. Particle Localization, 3. Tracking, 4. Refinement, 5. Numerical Reconstruction.

The design of the user interface (Figure 3B) follows the workflow shown in Figure 3A. Using the program interface, users can load frames from video streams (e.g., avi, mpeg, mov) and tune parameters to achieve optimal results. In the following, we describe available functions in the user interface and provide parameter values for data analysis (supplementary information section Main Interface).

*Pre-processing*: This step allows the user to improve contrast and to reduce noise in images using a background subtraction technique. More importantly, this step is required for DHM reconstruction since the hologram should be normalized by division with a background image. To generate a background image two options are available: "Static Background" creates a background image by averaging the intensity of fixed number of frames in a video. In contrast, "Moving Average" creates a constantly updated background image. By default, the number of frames used for the background image is set to 10 for a moving average filter. In this case, each pixel in the 11$^{th}$ frame is divided by the corresponding pixel in background image. The background image is then updated by replacing the 1$^{st}$ frame with the latest frame and averaging over all 10 past frames at each pixel.

We also provide options to change the image size, normalize the image intensity, and apply a Gaussian filter to remove noise in the image. Furthermore, the user can define a ROI for all video frames. This means, only objects in the ROI will be considered in the following "particle localization" process. Margins around this area are cropped. The size and position of the ROI can be selected manually.

*Particle localization*: The user can select two particle localization strategies: manual selection or auto-detection. For automatic detection, circle detection based on ITs transform is applied. The process starts with the edge detection of objects in the image. This is implemented by using a Canny edge detector [23], where the threshold value used in the detector is set manually by the user, or with the Otsu thresholding method [24]. The resulting binary edge image from the edge detector contains the information of edge positions. The ITs transform uses this information and generates a matrix with the same size as the original image, containing the probability distribution of circle centers. With the multi-scale solution (Figure 2A), the level of scales is by default set to 3 and we use the original image and two down sampled images with a factor of ½ and ¼ in both length and width. All three images are processed by using ITs transform, resized back to the original image size and fused into a final probability distribution of circle centers (Figure 2B). Thereafter, depending on the criteria to analyze the probability distribution, center positions can be extracted. A detailed description of these criteria and how they influence the particle center detection can be found in supplementary information (section Main Interface).

In addition to automatic detection, the program offers an option to manually select the center position of each particle. The center position from the manual selection is automatically refined using a variation of the cross-correlation method [25], which is based on the assumption of particle symmetry. Since the manual selection does not have detection results except for the 1$^{st}$ frame, the algorithm applies a local detection based on ITs transform for every selected object in the tracking process.

*Tracking:* For identified particles the program saves; particle positions, motion models, and template images. Based on the results from the localization process mentioned above, new particle positions from the detection can be assigned to previously identified particles. To ensure correct tracking we use the Hungarian method with a Kalman filter, which provides the optimum distance assignment [26,27]. One important parameter for the Hungarian method is the "Cost of NonAssignment" (Figure 3B, "3. Tracking"), which is set to 80. This means that the detection will not be assigned to identify particles if the distance between detection and predicted position is larger than 80 pixels. Furthermore, it is important to set the image template size in this step for DHM data because the template must contain information about background intensity and diffraction fringes. The user can set a fixed template radius to 200 indicating 401×401 templates. To ensure correct template values in the reconstruction process, the user activates a tracking boundary condition so that each template must have acquired all intensity values available in each frame.

*Refinement*: The center positions and templates derived from the tracking part of the program can be improved for better accuracy. The "Re-sampling" function creates a 1D profile of the particle in the radial dimension for axial numerical reconstruction as described in section 0. The program offers two sampling strategies. The option "Circular" (Figure 3B, "4. Refinement") samples in all radial directions, covering the full angular range. On the other hand, the "Sector" only samples pixel values in radial directions perpendicular to the moving direction of object to minimize the effect of motion blur in the reconstruction process. Besides, the user can also add a cross-correlation based module "Xcorr" [25] to refine the trajectory of each particle by finding the geometrically symmetric center position with sub-pixel accuracy.

*Numerical reconstruction*: In this module, intensity reconstruction in the axial direction is performed (Figure 1, right). By choosing option "1D" (Figure 3B, "5. Numerical Reconstruction"), axial reconstruction is accomplished with the Rayleigh-Sommerfeld model presented in section 0 that uses only a 1D radial intensity profile. This intensity profile is sampled from each particle template. Then the "Template Margins" are set to 2× the original profile size to reduce artifacts during the calculation of hologram reconstruction. This new radial intensity profile is extended on both sides by extrapolating averaged intensity values. Thereafter, the initial step, step size and last step is defined in micro-meter along the axial direction for position reconstruction. The reconstructed distance is calculated by determining the distance that has the maximum reconstructed intensity along profile center. To reduce errors from incorrect particle center identification, the user can apply a mask around the particle center within a range defined by the "Paraxial mask radius" to collect averaged reconstructed intensities. Moreover, the program provides the "1D Decov" option activating the deconvolution process [28]. Using this option, the reconstruction of a point source is deconvolved with the reconstruction results from a hologram so that a better signal to noise ratio can be achieved. As an alternative to the fast one dimensional approaches, we also offer a "2D" choice which is based on the implementation of the angular spectrum method using Equation (1) and using the intensity profile from the template. This option can be used to verify the results from other implementations, but it is computationally expensive.

The program also offers several useful functions to visualize tracking results. The user can activate "Present Tracking Images" to see if objects are correctly tracked. The user can select a single detected object manually and visualize the intensity reconstruction in the axial direction by activating "Show Intensity Reconstruction". The program will automatically save the tracking data into a user-defined folder. Additionally, used program configurations can be saved.

The post-processing interface (supplementary information section Post-Processing Interface) allows the user to visualize results either from a single trajectory file ("Single Trajectory") or from a group of trajectories in a folder. The option "Flow Profiling" collects trajectories of all particles identified in the tracking process and plots them in a 3D space as a color map with colors indicating the velocity. As an alternative, the average speed versus average height can be plotted for each trajectory. These data can be fitted to a polynomial function to quantify the particles speed at different height in the flow chamber. Additionally, users can set the reconstruction axial distance range in μm to present trajectories only within selected height.

# 4. RESULT AND DISCUSSION

## 4.1. 2D Particle Tracking: Center Detection with Nanometer Accuracy

We validate the 2D tracking performance of UmUTracker using the workflow described in section 3.3 with the axial reconstruction process deactivated. In this case, we determine only the lateral particle positions from images acquired using both bright field and digital holography microscopy (section 2.2).

First, we track the position of a single supercoiled DNA tethered particle under bright field microscopy conditions, as shown in Figure 4A. The raw data are taken from [16,17]. To determine the center position of the particle from a time series of images, we run UmUTracker in its standard configuration without background subtraction (Figure 4A, blue data points). As it is expected for restricted Brownian motion, we find a Gaussian distributed movement for the tethered particle (Figure 4A, histograms). This conclusion is in agreement with the results published in [16,17].

We further test the 2D multiple tracking algorithm by selecting 30 samples of 1 µm particles which are immobilized on a coverslip (Figure 4B, top images), illuminated with the LED. The piezo-stage is driven with an oscillation amplitude of 250.0 nm and a frequency of 0.50 Hz. Videos of the measurement are acquired at 200 Hz and can be downloaded from the supplementary material [29]. To analyze the experimental data, we run UmUTracker in its default configuration, but adapt the threshold value for edge detection to ensure stable detection and tracking of particles in the recorded video. By analyzing the oscillation data in the $y$ direction, we retrieve the expected particle motion (Figure 4B, black data). A sine function is fitted to the data providing an amplitude of $252.0 \pm 0.5$ nm and a periodicity of $0.50 \pm 0.01$ Hz (Figure 4B, blue line). Therefore, the difference between measured and fitted amplitude and periodicity is less than 1%. We observe comparable results for oscillations in $x$ direction (data not shown).

From the experiments above, we demonstrate that UmUTracker can automatically detect particle motions in the $xy$-plane in two different applications. The tracking accuracy depends on the microscopy system, for example, the optical zoom can be applied to increase magnification so that the accuracy can be improved. The parameters for localization and tracking in UmUTracker need also to be tuned to balance between performance and accuracy.

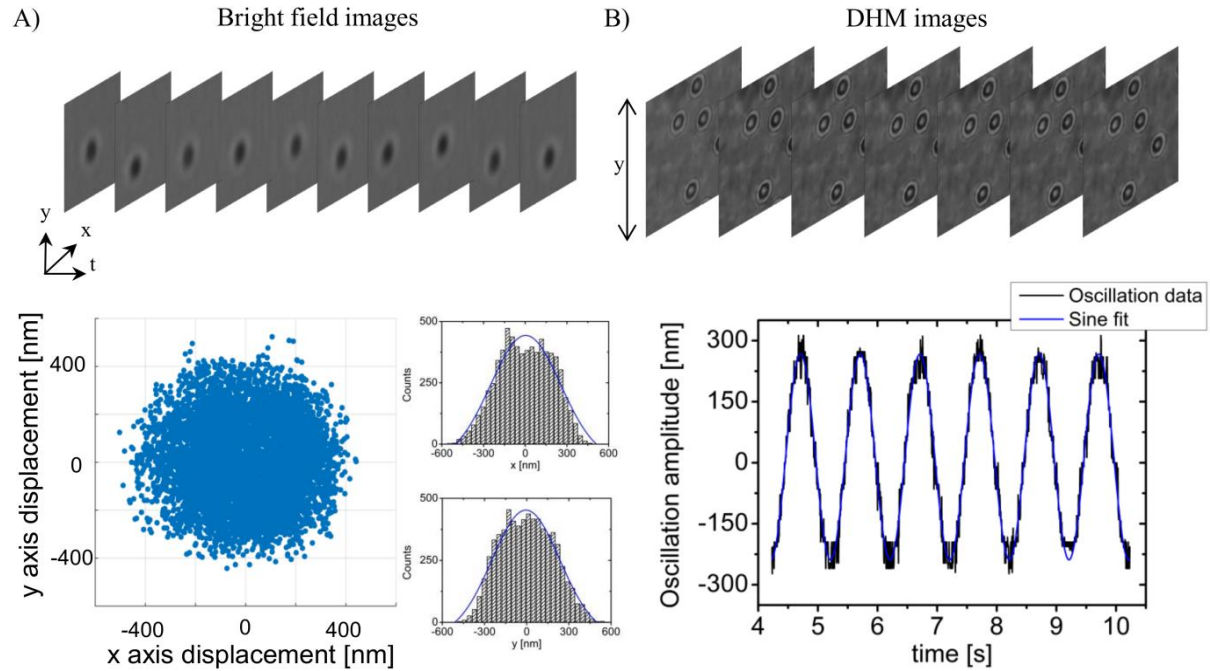

Figure 4 A) Experiments using bright field experimental data of supercoiled tethered DNA undergoing restricted diffusion. Gaussian fitting is applied to the displacement distribution in $x$ ($R^2 = 0.94$) and $y$ ($R^2 = 0.96$) direction. B) Sinusoidal oscillation of a 1 µm PS particle in $y$ direction (black data) with preset oscillation amplitude of 250 nm and a periodicity of 0.5 Hz determined by UmUTracker. The fitted oscillation amplitude and periodicity (blue line) differ less than 1% from the values set for the piezo stage movement. This agreement is confirmed by a coefficient of determination $R^2 = 0.98$.

## 4.2. 3D Particle Tracking: Numerical Reconstruction of the Axial Particle Position with Micrometer Accuracy

To determine the axial particle position, first we conduct a simulation using OpticsStudio16.5 software (Zemax LLC), producing synthetic diffraction patterns of PS particles without background noise. In the simulations, 10 spherical objects with refractive index of 1.50 and radius of 3.50 µm are placed in water with refractive index of 1.338 at reference distances 40, 50, 60, 70, 80, 90, 100, 110, 120 and 130 µm, from a virtual detector (Figure 5A). A $300 \times 300$ µm$^2$ virtual detector in OpticsStudio counts the number of rays passing through a surface without affecting the propagating rays. Rays are generated from a collimated light source of 470 nm wavelength and 4 billion rays are traced in a simulation. The image generated by the virtual detector has a size of $1500 \times 1500$ pixels resulting in a 200 nm/pixel conversion factor for reconstruction. Tracking of particles is performed by using the simulation data including 100 frames of the same synthetic image. Some key parameters used in UmUTracker for ITs transform to detect particle centers are: $401 \times 401$ pixels as the region-of-interest for each particle, 3 layers in scale, ITs threshold of 0.03 and the iteration of randomized sampling is set to 10× the number of binary edge points. We find a linear relationship between the reference positions and reconstructed distances from the focal distance of the particle to virtual detector (Figure 5B). In this context, it should be clarified that a transparent particle acts as a lens, by focusing incoming light at its focal distance. Using the Rayleigh-Sommerfeld back propagation theory we calculate the position of this focal point, to which we refer as "reconstructed distance". The reconstruction uncertainty is ~ 40 nm. This indicates that reconstruction of particle heights using UmUTracker is accurate for images acquired under decent light conditions and without background noise.

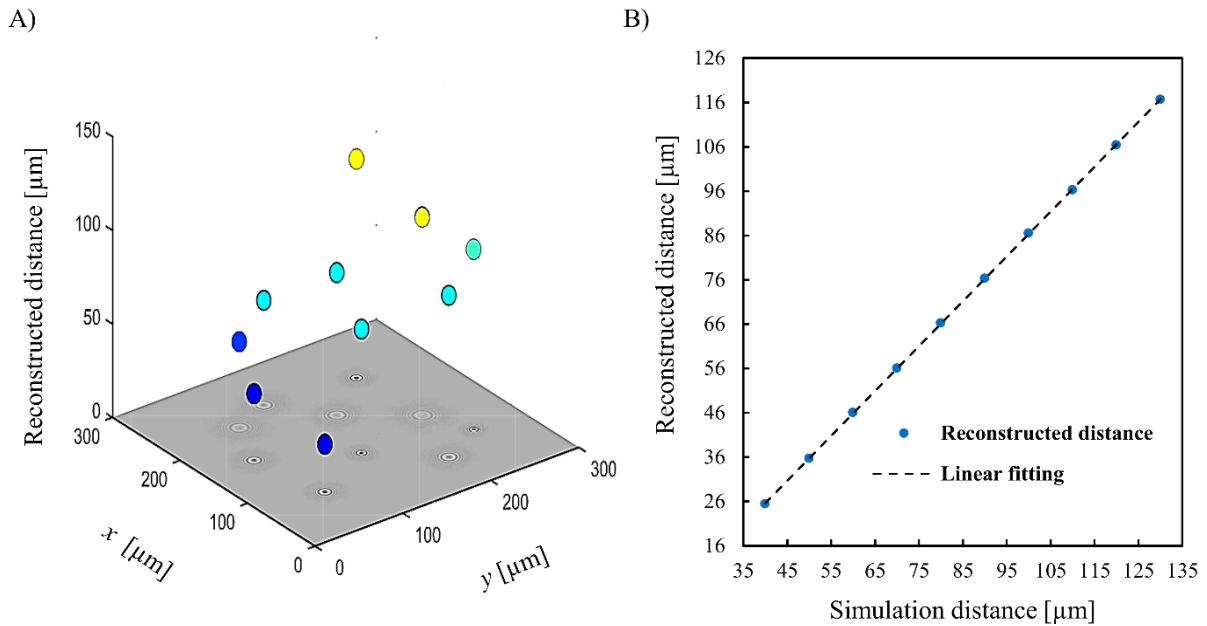

Figure 5 Relation between numerical reconstruction of particle intensity and estimated position in $z$ axis using the synthetic data from OpticsStudio16.5 software simulation. A) presents the reconstructed distance in 3D space. B) presents the linear relationship between simulated $z$ distance to the detector and the reconstructed distance (dashed line). Error bars at each point of reconstructed distance represents the standard deviation of measurement using UmUTracker averaged over 100 frames. The reconstruction uncertainty is approximately 40 nm. The coefficient of determination $R^2 = 0.99$ indicates the fitting of linear model with a slope equal to 1.0 is accurate and therefore the reconstructed distance coincides with the theoretical heights.

We also conduct experiments using 1 µm PS particles immobilized on the bottom coverslip of the measurement chamber, as explained in section 2.1. In detail, we first focus on the stationary particles on the coverslip. Then we perform a 1D scan along the *z* axis, covering a range between -50 µm to 50 µm (Figure 6A) around the immobilized particles on the coverslip with a step size of 1 µm. At each position 40 frames are acquired at a frame rate of 200 Hz (Figure 6B, spherical data points).

In a range from 2 µm to 50 µm UmUTracker provides a linear relationship between piezo stage movement and reconstructed distance (Figure 6B, blue data points). Based on the slope of the linear fit the reconstruction underestimates the 1 µm steps by approximately 1%. The reconstruction uncertainty is given by the standard deviation at each step and the averaged value is calculated to be ~150 nm. Around the focal region ranging from 0 µm to 2 µm (Figure 6A, middle and Figure 6B), the reconstruction cannot, however, resolve the 1 µm steps, but shows a plateau around $0.8 \pm 0.9$ µm. This can be explained by the fact that both the method based on reconstructing intensity and the theory of Rayleigh-Sommerfeld are not derived to handle such situations. Interestingly, the reconstruction of the particle height gives reasonable results even if the hologram is produced by the interference between the non-scattered and the backscattered light from the particle (Figure 6B, red data points). In this scenario, the image plane is located between the light source and particles (Figure 6A, top), in contrast to a classical in-line holography alignment, which is: light source, particles and image plane (Figure 6A, bottom). Nonetheless, we observe again a linear relation between reconstructed focal distance and piezo stage movement, covering a range from -3 µm to -40 µm (Figure 6B, yellow line). Below -40 µm focal distance the reconstruction algorithm becomes error-prone due to blurry diffraction pattern in the image. This applies also to focal distances above 50 µm (section 4.3, Figure 6B).

To improve the reconstruction accuracy in these two extreme cases, the illumination can be optimized by using low coherent laser illumination [30]. Furthermore, the CCD sensor can induce measurement bias [31] and noise generated from thermal electrons within the exposure time can produce diffusive diffraction patterns. This effect becomes only prominent if short shutter times are used to acquire images. To compensate for this and produce higher definition holograms a more sensitive sensor, higher light intensities, and an objective with higher magnification should be used. Moreover, by using a microscope objective to record images with a CCD camera, the objective lens add a phase shift that impair the axial intensity reconstruction accuracy, especially for objects far away from the image plane [32]. Despite these deviations, we show that UmUTracker can reconstruct the distance of a particle to the image plane over a height range of 50 µm with high accuracy.

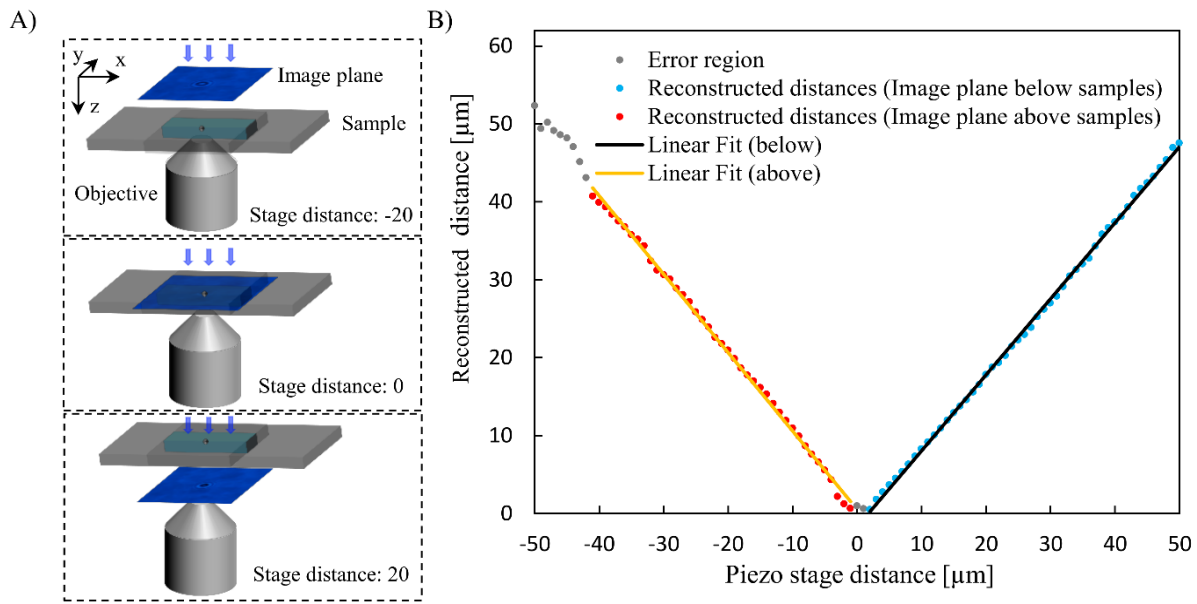

Figure 6 A) Various positions of the sample during the 1D scan in the axial direction. In the top image, the sample is positioned between the objective and image plane. During the scan, the sample is moved through the image plane (middle), approaching its final position with the image plane located between sample and objective (bottom). The latter scenario corresponds to the classical in-line digital holography. B) Reconstruction of the focal distance for 1 µm PS particles from experimental data acquired at 200 Hz. The reconstructed focal distance (blue data points) shows a linear relationship (black line, slope = 0.99, $R^2 = 0.99$) with the piezo stage distance. At each step, 40 images are recorded resulting in a standard

deviation of 150 µm. Even for a scenario in which the image plane is positioned between light source and particle, UmUTracker reconstructs the particle height correctly (red data points, yellow line) showing again a linear behavior (slope = -1.03, $R^2$ = 0.99).

## 4.3. 4D Particle tracking: Characterizing the Flow Profile in a Microfluidic Chamber

After verifying the accuracy of UmUTracker for 2D particle tracking and 3D axial position reconstruction, we now analyze the flow profile in a rectangular microfluidic chamber by tracking the velocity of 1 µm particles moving at different heights in the flow. For that purpose, we use the commercial available 100 µm high microfluidic bio-chip (section 2.1) and create a constant and reproducible flow rate of 26 ± 2 nL/s in the $x$-direction using a microfluidic pump equipped with a 10 µl pipette. A 150x150 µm area in the $xy$-plane in the central part of the chamber is imaged ~5 µm below the bottom surface with a camera frame rate of 200 Hz. We set the UmUTracker settings according to the following; a 401 × 401 pixels region-of-interest for each particle, 4 layers in scale, ITs threshold of 0.1 and the iteration of randomized sampling is set to 5× the number of binary edge points. To visualize the flow cross section perpendicular to the flow direction, we collect all trajectories of tracked particles and calculate their mean position in all directions $x,y,z$ (Figure 7A). The speed along the $z$ axis (reconstructed distance) of this cross section is calculated by taking the distance between the first and last position divided by the travelling time for each trajectory (Figure 7B).

For a flow rate of 26 nL/s the flow velocity ranges from 10 µm/s to 1050 µm/s for particles close to the bottom cover slip and particles moving in the center of the flow chamber (Figure 7A, blue and red data points). Fitting the velocity profile along the $z$ axis of the flow cross section with a polynomial of second order, confirms its parabolic shape as expected for a laminar Poiseuille flow (Figure 7B, dashed red line). In this context, it must be mention that only data points below a reconstructed distance of 55 µm are considered for fitting (Figure 7B, blue data points), since above this height we observe a pronounced uncertainty in our measurement results. This can be explained by the fact that diffraction patterns for particles far away from the image plane produce only diffusive diffraction patterns, reducing the accuracy of the reconstruction routine (section 4.2). Additionally, the diffraction pattern of fast moving particles in the flow center are blurred and extended in the moving direction. We will discuss the influence of motion blur on our results in the following paragraph.

To verify our flow data and its velocity profile, we simulate the latter using CFD simulations. Since the experimentally determined flow rate is 26 ± 2 nL/s, we conduct the simulations for a mean flow rate of 26 nL/s and for the two extreme values, that are 24 nL/s and 28 nL/s (Figure 7B, light blue line and grey area). With a maximum flow velocity of 1150 mm/s the simulation result for 26 nL/s overestimates the experimental value by 3 %. Therefore, our flow data are better described by simulations assuming a flow rate of 24 nL/s. Furthermore, the experimentally reconstructed flow chamber height exceeds the specifications from the supplier by approximately 5%. Both deviations in maximum flow velocity and chamber height can be attributed first to a slight misalignment of coordinates between sample stage and camera during the measurement (Figure 1, right). Second, due to the motion blur the algorithm overestimates reconstructed distances in axial direction for fast moving particles and for those far away from the image plane. In contrast, the flow velocity error originating from poor particle center detection in lateral plane is comparably small, as it turned out that the tracking part of UmUTracker is robust, providing nanometer resolution even under noisy conditions (section 4.1). To further improve the detection accuracy against motion blur, a refinement algorithm, such as Xcorr or C-Sym [33], can be applied. Despite these deviations, our result shows the correct flow profile in a microfluidic channel over a height range of approximately 55 µm, exceeding the depth of field of 60x/1.40 objective by a factor above 100. Furthermore, we received decent agreement between measured and simulated flow profile, proving that UmUTracker characterize microfluidic flows with high accuracy.

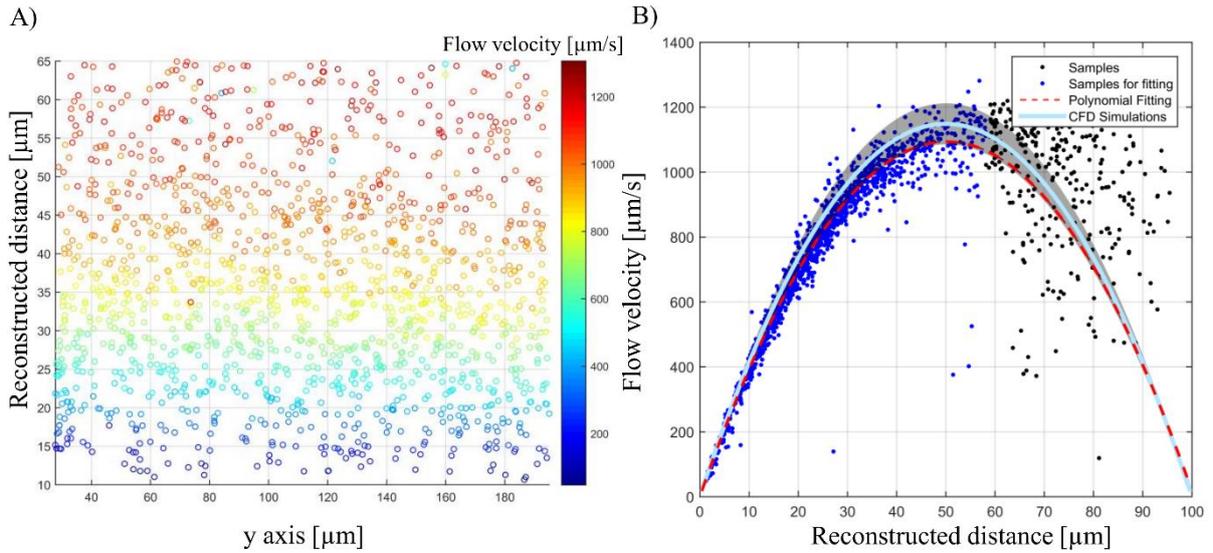

Figure 7 Microfluidic flow profiling using UmUTracker. A) presents sampling coordinates at averaged reconstruct distance (axial direction) and averaged *y* axis positions (horizontal direction which is not the flow direction) for each of the trajectory. The color code represents the averaged flow velocity at different heights. B) presents the flow profile in axial direction. Samples of averaged velocity at different axial positions are marked as dots regardless of color. A second order polynomial function is fitted to the samples ($R^2 = 0.93$) in blue as dashed red line and the simulation results using CFD is presented in light blue line with error margins in grey.

### 4.4. Execution time

To evaluate the overall performance of UmUTracker, we conducted tracking experiments using a computer with an Intel® Core™ CPU i7-4770 processor and 16 GB of RAM memory running Windows 7 professional. The UmUTracker used template size $201 \times 201$ in a $1024 \times 1024$ image for tracking and axial intensity reconstruction with a speed of 5 fps. Comparing with the 2D reconstruction with parallel computing utilizing 4 CPU cores based on Equation (1), our 1D version of reconstruction without any parallel computing can provide results with high accuracy and with performance more than 10 times faster depending on the size of the template.

## 5. CONCLUSION

We present an automated software for tracking particles using images acquired with bright field or digital holographic microscopy. A robust circle detection algorithm is implemented that is based on the ITs transform for localizing diffraction patterns and a resource-efficient solution for numerical reconstruction of spherical particles using 1D profiles. The software is capable of handling 2D tracking with nanometer accuracy, and 3D tracking for an in-line DHM setup as well as flow profile estimation as velocimetry using micro-meter sized PS particles. Reliable automatic tracking makes flow measurements much simpler and the execution time with a $201 \times 201$ template in a $1024 \times 1024$ image can reach 5 fps. Due to the object-oriented programing style, the software is easy to modify with customized modules. The application of this program is not limited to particle tracking or flow measurement. It can also be used for tracking of cells and bacteria. The latest version of the program is available here: https://sourceforge.net/projects/umutracker/

## ACKNOWLEDGEMENTS

T.S. acknowledges financial support from the German Research Foundation (DFG) via a postdoctoral fellowship. This work was supported by the Swedish Research Council (2013-5379) and from the Kempe foundation to M.A.

# Supplementary Materials

UmUTracker: A versatile MATLAB program for automated particle tracking of 2D light microscopy or 3D digital holography data

Hanqing Zhang[†], Tim Stangner[†], Krister Wiklund[†], Alvaro Rodriguez[†], Magnus Andersson[†]

[†]Department of Physics, Umeå University, 901 87 Umeå, Sweden

## Table of contents

# Getting started

### 1. Requirements

UmUTracker is developed in MATLAB (version R2014b) using the image processing toolbox to process video frames. UmUTracker is tested on Windows 7 and 10 operating a 64-bit system. We recommend a minimum of 4 GB of RAM memory and sufficient hard drive space to ensure reliable operation when running long video files. A 2.0+ GHz processor or higher is recommended to reduce the execution time.

### 2. Video file preparation

UmUTracker can process video files, supporting a wide variety of video types; .avi, .mp4, .m4v, .mov. We recommend to acquire sample videos with highest quality to ensure reliable object identification by the tracking algorithm. To avoid motion blur, make sure that a video is captured using a proper frame rate and shutter speed.

### 3. Load Video

Before loading the video, please check the default tracking parameters of each function and tune them if necessary. In general, the tracking functions are disabled before loading the video. To load a video, please press '*File*' in the menu options and then press *'Load'*, or simply press *'Load Video'* (see Figure S1) and select a sample video. Once the video is loaded correctly, the 'Input path' and 'File name' is automatically filled in with corresponding information of the sample video. The '*Check Edge Detection*' as well as '*Start*' buttons will now be activated.

# Quick start

## Basic Procedures

1. Unpack all files from UmUTracker.zip to a folder. Double click the UMUTracker.m and open MATLAB. Run UMUTracker.m in the MATLAB to load the main interface.

2. Load a video and then choose between '*Manual*' or '*ITs Transform*' in the panel group '*Particle Localization*'. Thereafter press *'Check Edge Detection'* to analyze the detection results. This can require some tuning depending on the video quality. For 2D tracking, untick the checkbox '*Enable Numerical Reconstruction*'. All parameters for tracking can now be changed, saved or load from a .mat file.

3. Click '*Start*' for tracking. If '*Manual*' is selected, click the particle center in the image and double click the last particle center to enable tracking. You can pause or stop the tracking by pressing 'Pause' or 'Stop' button.

4. For optimal tracking, you need to tune the parameter values for the different functions. Below, we explain in details how this affects the data and give some suggestion when to use what values.

## Demo video example

We provide a demo video 'DemoVideo.avi' of a micron-sized particle recorded using in-line DHM. The bead is in water (n = 1.33) and attached to the bottom of coverslip. We use light at 470 nm to illuminate the sample and pixel to nm conversion factor is 132 nm.

We provide a configuration file 'DemoVideoConfig.mat', which includes the parameters used when we analyzed the video. This configuration file is loaded by pressing the *'Load'* button under the configuration panel in the main interface.

# Main Interface

The main interface controls the file input-output functionalities and all tracking features can be tuned.

## Major functions

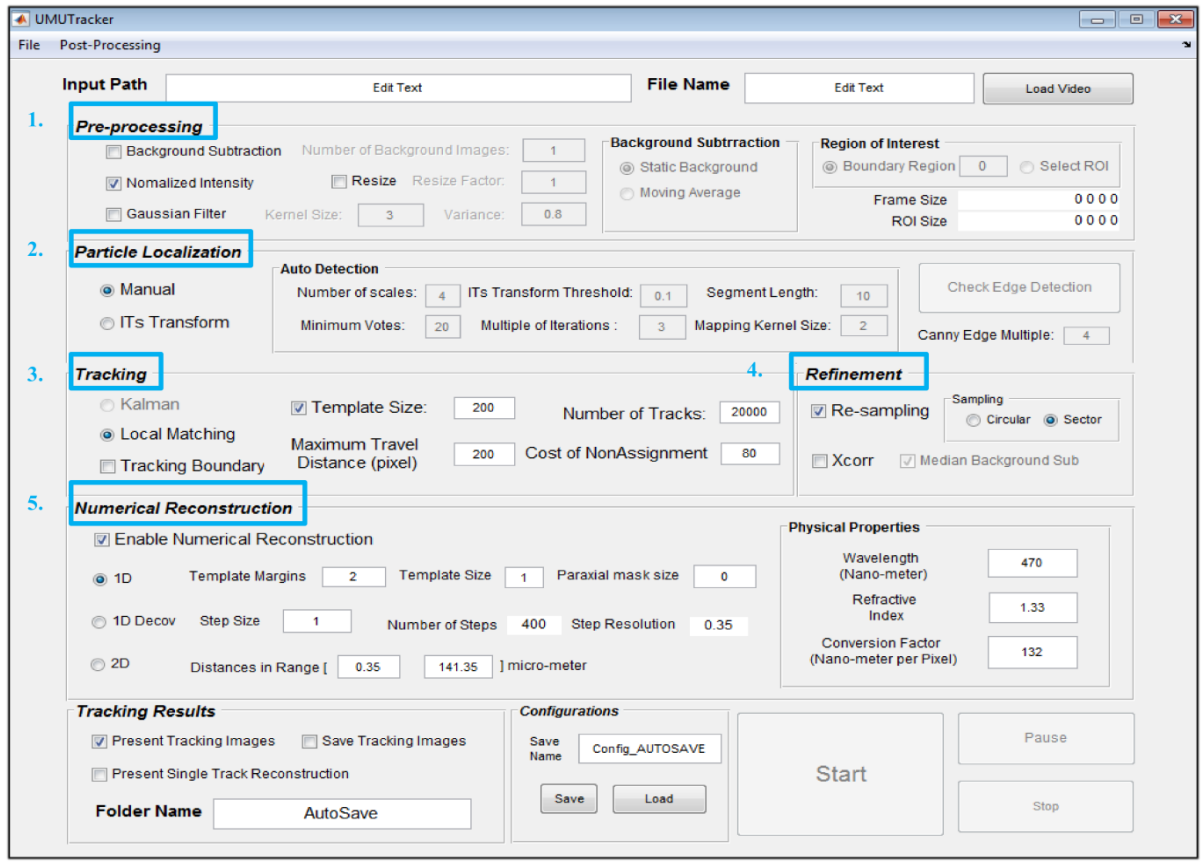

Figure S1. Overview of the main interface. Five major functionalities are presented and marked as: 1. Preprocessing, 2. Particle Localization, 3. Tracking, 4. Refinement, 5. Numerical Reconstruction.

### 1. Pre-processing

To activate background subtraction, select the function '*Background Subtraction*' and set the '*Number of Background Images*' used to create a background image. This value is set to 1 by default. The background image is updated in two ways. '*Static Background*' creates the averaged image by using the number of frames set, starting from the 1st frame of the video. '*Moving Average*' creates the averaged background image by using the current frame and previous frames set by '*Number of Background Images*'.

The '*Region of Interest*' function contains two options. The '*Boundary Region*' option allows you to set the ROI for all video frames as an image cropped from the original frame by removing pixels close to the edge. The number of pixels removed is set to 0 by default. Alternatively, the ROI can be selected manually with '*Select ROI*'. In this case, the 1st frame of the video will be shown in a separate window and you can drag a squared region and double click to activate the ROI region. After the ROI is selected the ROI size will be updated and presented along with the frame size.

The '*Normalized Intensity*' function sets the image intensity range. The range is [0,1].

To change the original frame size, the '*Resize*' option can be used by changing a factor set by '*Resize factor*'. By default this value is set to 1.

To remove noises in the image, a Gaussian filter can be used, '*Gaussian Filter*'. The size of Gaussian kernel in pixels can be set with '*Kernel Size*'.

**2. Particle localization**

Two particle localization strategies can be selected: '*Manual Selection*' or '*ITs Transform*'.

The '*Manual Selection*' activates the '*Local Matching*' in the '*Tracking*' process. When you click the '*Start*' button for tracking, the first frame of the video will pop-up and you can single click the object in the image to be tracked and double click the last object when all objects are labeled.

The '*ITs Transform*' activates the '*Auto Detection*' option. The '*Number of Scales*' defines the number of images at different Gaussian scale space. This number is related to the size of the image and a reasonable value is in the range [1,4]. A larger value is suitable to detect large diffraction patterns. The '*ITs Transform Threshold*' defines the minimum geometrical criterion value to identify a circular shape based on references [1]. This value can be set in range [0.01,0.3]. A smaller value defines circles with more accurate geometrical shape, resulting in a fewer number of valid candidates and less detected false-positive. A large value increases the number of detections but with a possibility to increase false-positive detections. The '*Segment Length*' is set to 10 indicating that at least 10 connected edge pixels are need for circle detection. Short segments and discontinued edge points are considered as noise and can be removed from the detection for efficiency. A smaller number means more detections with increasing false-positives. This value is preferably set from 3 to 100. The '*Minimum Votes* ' sets the threshold for selecting optimal detections. This value is set to 30 by default indicating the diffraction pattern center is recognized only when the voting number at a pixel is above this value. This value depends on the radius and completeness of circular pattern and a practical value is in range [5,50]. Results with less false-positive can also be achieved by changing "Minimum Votes" to a higher number. The '*Multiple of Iterations*' sets the number of edge pixels to be checked for circular patterns using randomized sampling. This value is set to 5× the number of edge pixels by default and can be change from 1 to 10. A large value increases the number of detections and improves the precision of detection in the image but with longer execution time. The '*Mapping Kernel Size*' represents the radius of vote area. This value is set in range [1,10] and a large value increases detection accuracy for object with large radius. When this value is set to 2, a 5×5 Gaussian kernel is superposed in the distribution of votes.

The '*Check Edge Detection*' option provides an intuitive way to set the threshold value for edge detection. Once the button is pressed, a pop-up interface is presented, see Figure S2. This is important since Canny edge detection [2] is applied as a pre-requisite for the ITs transform. We apply the Otsu thresholding method [3] to get a default threshold value. The resulting value is applied directly to Canny detection as higher bound of threshold, and a lower bound is set to one-third of this value. In practice, users can set the higher bound manually by changing the *'Canny Edge Multiple*' within [0,10].

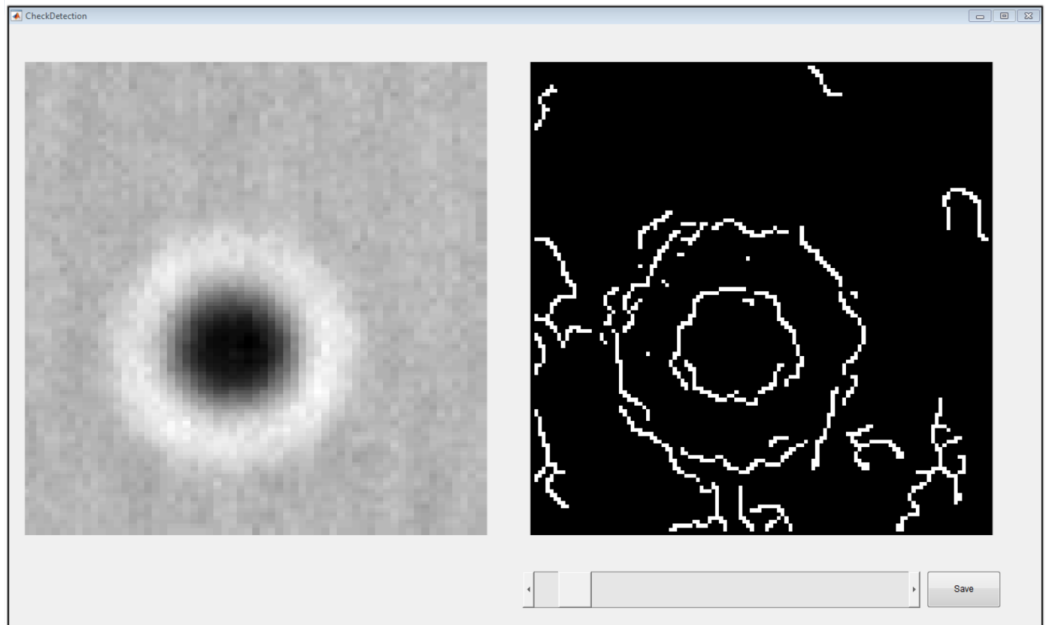

Figure S2. Interface for checking the edge detection. The original image frame (left) and edge extraction results using Canny edge detection (right) are presented. The threshold value for edge detection can be set by dragging the slider and save the corresponding values by pressing the button '*Save'*.

## 3. Tracking

The option '*Kalman*' ensure improved tracking using the Hungarian method with a Kalman filter, which provides the optimum distance assignment [5,6]. As an alternative, the option '*Local Matching*' can be selected so that a local matching using ITs transform is applied to image templates to find the diffraction center.

The parameter '*Cost of NonAssignment*' for Hungarian method is set to 100, meaning that the detection algorithm will not identify particles if the distance between detection and predicted position is larger than 100 pixels. Similarly the '*Maximum Travel Distance*' set the upper bound of distance change regardless of the tracking method.

The '*Number of Tracks*' set the maximum number objects to be tracked in the program.

The '*Template Size*' is set to 200 indicating a 401×401 template is created around the particle center. This value can range from 10 to 500 depending on the size of object in the video frame. To ensure correct template data for axial reconstruction process, the user can activate '*Tracking Boundary*', meaning that each template must have acquired all its intensity values within the range of frame size.

In the case that no detection results are available when 'Manual Selection' in the localization process is chosen, a local detection using ITs transform is applied to image templates to find the diffraction center.

## 4. Refinement

The '*Re-sampling*' function creates a 1D profile, which can be set using either one of the two following sampling strategies. The option '*Circular*' samples in all the radial directions, whereas the '*Sector*' only samples pixel values in the direction of the moving object to minimize the effect of motion blur. Thus, we recommend to use '*Circular*' when detected objects are not exposed to motion blur.

The '*Xcorr*' modules [6] refines the position of each track using the intensity profile and a fixed sized template of the object to find the geometrical center position with sub-pixel accuracy. This, however, is a computational heavy process. *'Median Background sub'* improves the accuracy of the algorithm when the median value in the image represents the intensity of the background pixels.

**5. Numerical reconstruction**

If experimental data are acquired using digital holographic microscopy (DHM) and the spatial position ($xyz$) should be determined, please activate *'Enable Numerical Reconstruction'* to estimate the axial object position.

Set the '*Wavelength*' of the light used for illumination, the '*Refractive index*' of the solution, and the '*Conversion factor*' between pixel and (nano-)meter scale in the $xy$-plane. Since the accuracy of the reconstruction depends strongly on experimental conditions, properties such as the wavelength of the light used for illumination, refractive index of solution, and conversion factor must be set or measured correctly.

The program offers different reconstruction approaches. The option '*1D*' can be selected if users want to use the angular spectrum method with a 1D radial intensity profile. The '*1D Decov*' option use a deconvolution process [7], and the '*2D*' option use an implementation of the 2D angular spectrum method [8].

The '*Template Margins*' for reconstruction is set 2× the size of intensity profile to avoid artifacts during the reconstruction process. The '*Template size*' can also be changed, but please keep in mind that this is a trade-off between efficiency and accuracy. This value is by default 1 meaning the template has its original size defined in the '*Tracking*' process. If a badly derived reconstruction is made due to a template with its particle center calculated with some error, the user can apply a mask around the particle center within a range defined by the '*Paraxial mask radius*' to collect averaged reconstructed intensities. Reconstructed intensities are sampled at each '*Step Size*' in the axial direction with the unit of micro-meter. Users can change the reconstruction distance to adjust the applied range of the numerical reconstruction. The '*Number of Steps*' and '*Step Resolution*' are automatically calculated according to the range of distance and step size.

## Tracking Results

Users can activate '*Present Tracking Images*'. By doing so, tracked objects are tagged with their identity number, tracking score, axial depth position and speed of the particle calculated using the current and previous positions (see Figure S3A).

The '*Save Tracking Images*' saves the trajectories of an object to a .txt file in a folder with its name defined by '*Folder Name*'.

One can also select a single target manually and visualize the intensity reconstruction in the axial direction by activating *'Present Single Track Reconstruction'* (See Figure S3B).

## Configurations

In '*Configurations*', parameters for different experiments can be saved by clicking '*Save*' with its file name written in 'Save Name'. The parameters are stored in a .mat format in the local directory of the program. Saved parameters can be loaded into the program by pressing '*Load*' button and select the corresponding .mat file.

A)

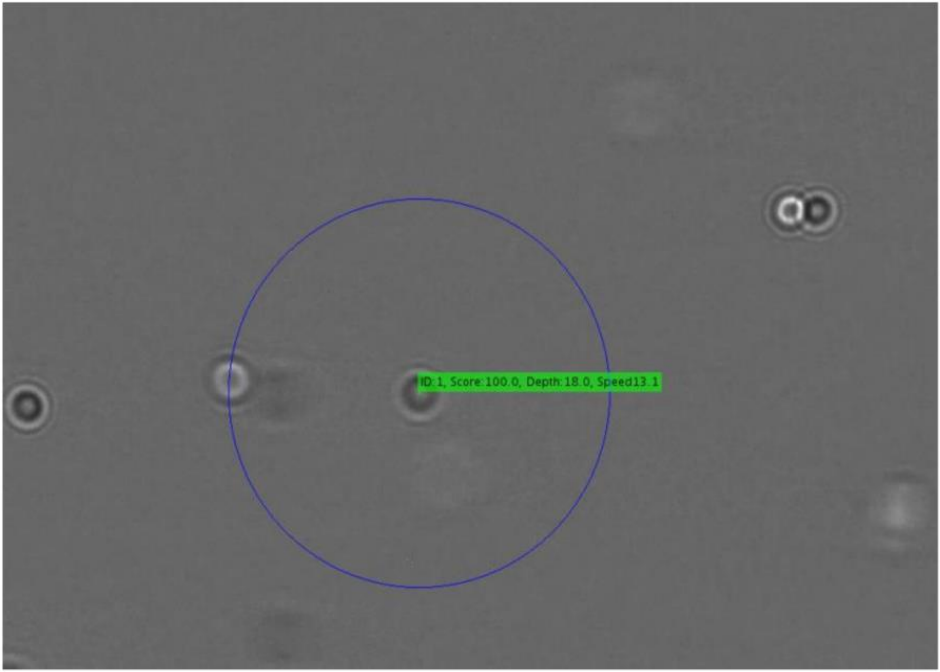

B)

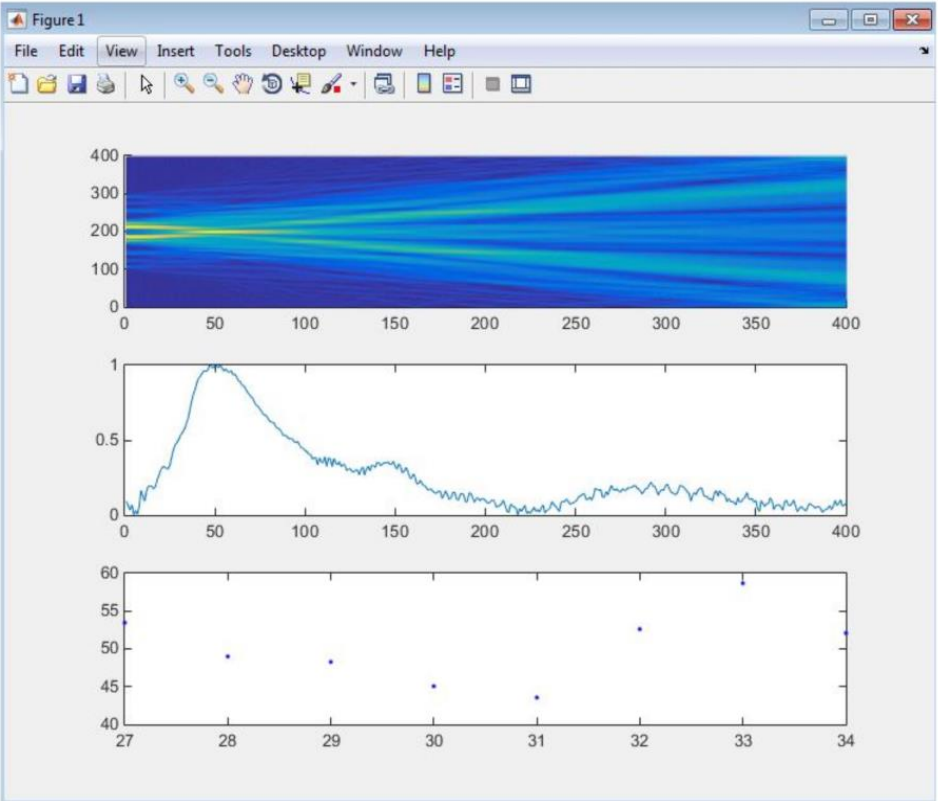

Figure S3 A) A demonstration of tracked object. The blue circle represents the size of the ROI for numerical reconstruction. B) The top graph represents the intensity profile reconstruction (vertical axis) at different axial distance (horizontal axis). The graph in the middle represents the axial intensity value along the center of the particle. The maximum intensity value (vertical axis) and the corresponding distance (horizontal axis) is then used to estimate the particle axial distance. The plot in the bottom shows a segment of the reconstructed distance in steps (vertical axis) vs. the frame number (horizontal axis).

# Post-processing Interface

We provide two post-processing interfaces, one to visualize the trajectory of a single object and one to analyze flow profile from multiple trajectories.

## Single Trajectory

The user can press '*Post-Processing*' in the menu options and then press *'Single Trajectory'*. The interface then provides the trajectory information in the *x, y* and *z* axis, and a 4D plot is created including the 3D trajectory, see Figure S4.

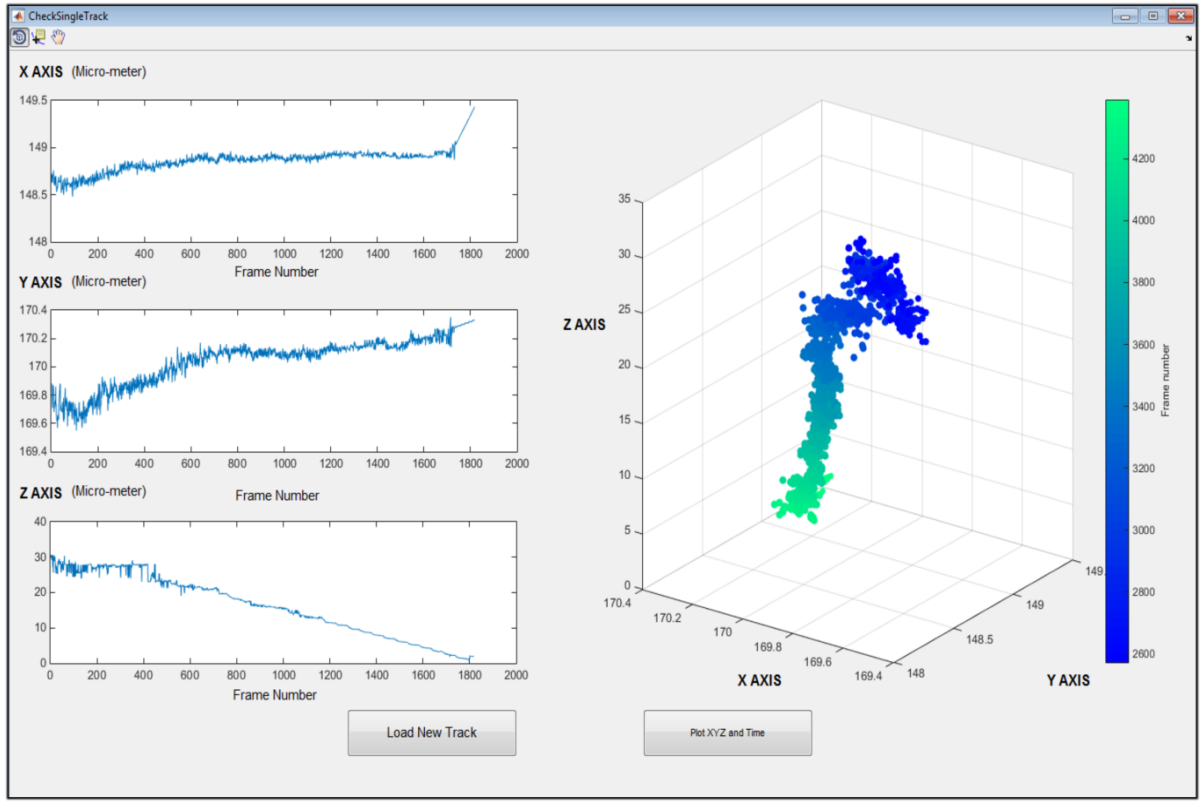

Figure S4. Interface showing the trajectory of a particle. The three plots on the left represent the frame number vs. object trajectory in *x, y* and *z* axis, respectively. The colored graph on the right shows the 3D trajectory with the time values colored according to the color bar to the right.

## Flow Profile

The '*Load Data*' option finds the folder of saved object trajectories. All saved .txt files in that folder are supposed to be created by UmUTracker during the tracking. After loading all the .txt data paths to the program, press '*Analyze*' to show the 4D flow profile. Then, a plot with the averaged axial position and the average speed is generated with a 2$^{nd}$ order polynomial fit, which is set as default. The same result is obtained by pressing '*4D Profile*'.

In '*Trajectory*', several criteria can be set to remove errors and outliers in the data. The '*Minimum Travel Distance'* sets the minimum difference of the first and last position detected in each object. The '*Minimum Track Size*' sets the minimum length of a trajectory. The '*Axial Standard Deviation*' sets the maximum value used for standard deviation in axial position. This value is to remove data points that changes drastically in axial direction, which can induce error during fitting. The '*Frame Rate*' is needed for calculating the flow speed profile. If the abovementioned conditions are not met, the corresponding trajectory is discarded for flow visualization.

To visualize the data, tune the option '*Sample Axial Distances*'. This uses the average trajectories from a given interval along the axial direction. An exemplary result is shown in Figure S5. Here, we set the option '*Sample Axial Distances*' to 5 and 65 micro-meter. Thus, only samples within this interval are plotted. Similarly, for fitting the sampled data, the user can set '*Curve Fitting Distances*' to choose the range for the fit. Besides, the *'Data Fitting'* options allow the user to set the *'Polynomial Order'* and the number of interpolated values based on the fitting function, *'Number of Fitting Bins'*.

The 'XZ view' and 'YZ view' allows the user to visualize the flow profile in the *x* or *y* direction. For example, if the fluidic flow is in *x* direction, the user can plot the *yz* view to check the 2D averaged flow profile.

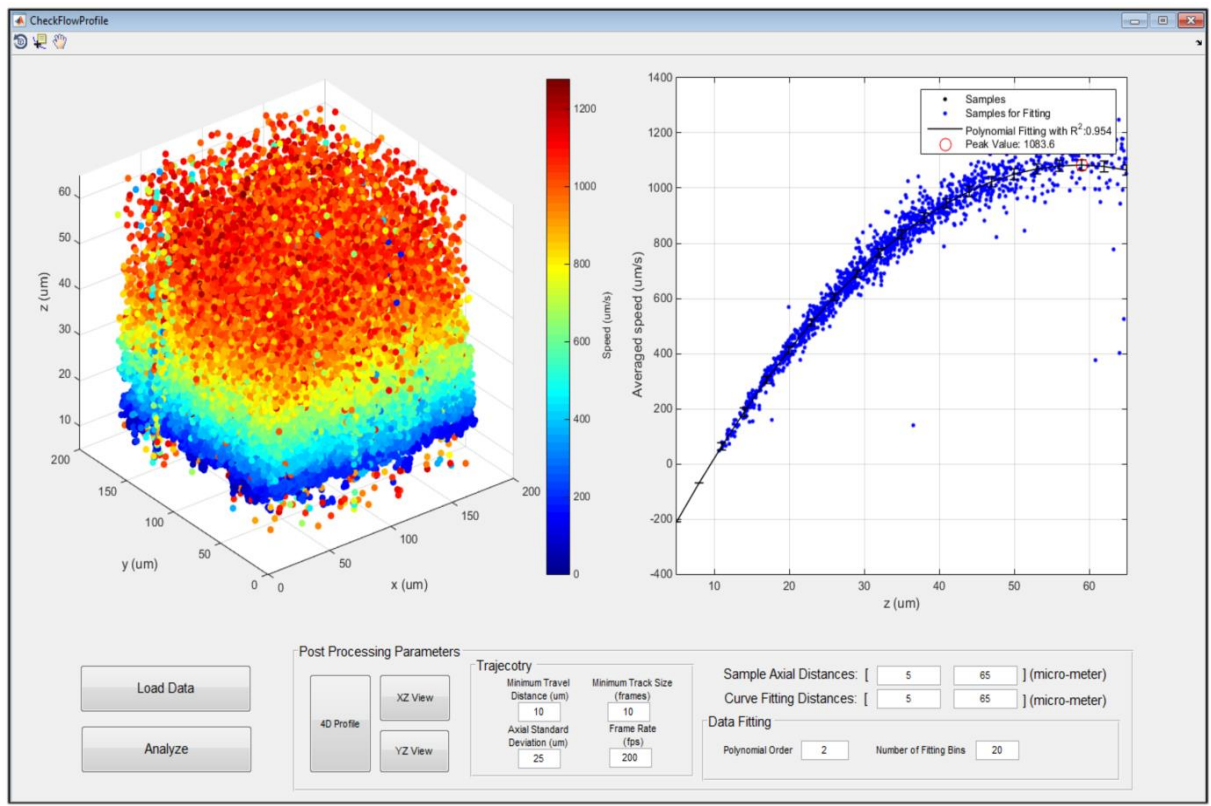

Figure S5 Interface for flow profiling.